%% file: arxiv-v2.tex
\documentclass[11pt]{article}

\usepackage[left=1in,top=1in,right=1in,bottom=1in,head=.1in]{geometry}

\usepackage{pratik}

\usepackage{./macros/macros_03_symbols}
\usepackage{./macros/macros_04_propernames}

\usepackage{booktabs}
\usepackage{multicol}
\usepackage{multirow}

\usepackage{adjustbox}

\newcommand{\HT}{\mathtt{HT}}
\newcommand{\DF}{\mathtt{DF}}

\newcommand{\SI}{\mathtt{SRS}}
\newcommand{\ST}{\mathtt{SSRS}}

\newcommand{\mse}{\mathrm{MSE}}

\newcommand{\auxmod}{\hat{Z}}

\newcommand{\hatm}{f}

\usepackage{xspace}
\newcommand{\kmeans}{$k$-means\xspace}

\usepackage[font=small,skip=5pt]{caption}

\usepackage[xspace]{ellipsis}
\renewcommand{\smash}{}

\date{}

\usepackage[
    textsize = scriptsize,
    bordercolor = White,
    textcolor =  Black,
    backgroundcolor = Yellow!10
]{todonotes}

\begin{document}

\newcommand{\titletext}{A Framework for Efficient Model Evaluation\\through Stratification, Sampling, and Estimation}

\setcounter{footnote}{1}

\newcommand{\footremember}[2]{%
    \footnote{#2}
    \newcounter{#1}
    \setcounter{#1}{\value{footnote}}%
}
\newcommand{\footrecall}[1]{%
    \footnotemark[\value{#1}]%
}

\author{
    Riccardo Fogliato\footremember{awsml}{Amazon Web Services; corresponding author email: \texttt{fogliato@amazon.com}} 
    \and
    Pratik Patil\footremember{berkeleystats}{University of California Berkeley} 
    \and
    Mathew Monfort\footrecall{awsml} 
    \and
    Pietro Perona\footrecall{awsml}~\footremember{caltech}{California Institute of Technology} 
}

\title{\titletext}

\maketitle

\begin{abstract}
\input{sections/abstract}
\end{abstract}

\input{sections/intro}
\input{sections/related_work}
\input{sections/framework}
\input{sections/design}
\input{sections/results}
\input{sections/discussion}

\bibliographystyle{splncs04}
\bibliography{bibliography}

\newpage

\appendix
\input{sections/appendix}

\end{document}

%% file: sections/abstract.tex
Model performance evaluation is a critical and expensive task in machine learning and computer vision. Without clear guidelines, practitioners often estimate model accuracy using a one-time completely random selection of the data. However, by employing tailored sampling and estimation strategies, one can obtain more precise estimates and reduce annotation costs. In this paper, we propose a statistical framework for model evaluation that includes stratification, sampling, and estimation components. We examine the statistical properties of each component and evaluate their efficiency (precision). One key result of our work is that stratification via \kmeans clustering based on accurate predictions of model performance yields efficient estimators. Our experiments on computer vision datasets show that this method consistently provides more precise accuracy estimates than the traditional simple random sampling, even with substantial efficiency gains of 10x. We also find that model-assisted estimators, which leverage predictions of model accuracy on the unlabeled portion of the dataset, are generally more efficient than the traditional estimates based solely on the labeled data.

%% file: sections/intro.tex
\section{Introduction}
\label{sec:intro}

Measuring the accuracy of computer vision (CV) algorithms is necessary to compare different approaches and to deploy systems responsibly. 
Yet, data labeling is expensive. 
While machine learning techniques are increasingly able to digest large {\em training} sets that are sparsely and noisily annotated, {\em test} sets require a greater level of care in their construction. 
First, the tolerance for annotation quality is much stricter, as annotation errors will lead to an incorrect estimation of model accuracy.  
Second, data must be collected and annotated at a scale such that
the confidence intervals around the error rates are sufficiently narrow (compared to the error rates) to make meaningful comparisons between models and error rates have been plummeting.
Lastly, for many applications, evaluating a single model can involve multiple test sets designed to assess performance in different domains, metrics, and scenarios.
This is necessary in testing, for example, cross-modal models such as CLIP \cite{radford2021learning}.
Practitioners facing the cost of putting together test sets will ask a simple question: \emph{How can one minimize the number of annotated test samples that are required to precisely estimate the predictive accuracy of a model?} 

Efficient estimation of model accuracy can be achieved by co-designing sampling strategies (for selecting which data points to label) and statistical estimation strategies (for calculating model performance).
One may craft sampling strategies that maximize the (statistical) efficiency of a given method for estimating model accuracy, that is, minimize its error given a fixed number of annotated samples \cite{imberg2022active,kossen2021active}.
Unlike simple random sampling, which picks any example from the dataset with equal probability, efficient strategies will select the most informative instances to annotate when constructing a test set.
One may also look for efficient estimators. 
Unlike design-based approaches, which base the statistical inference solely on the labeled sample, model-assisted estimators leverage the predicted labels on the remaining data to increase the precision of the estimates \cite{breidt2017model,wu2001model,sarndal2007calibration}.
However, CV researchers continue to rely on simple random sampling and design-based inference. 
Why?

We believe that there are two reasons why efficient sampling strategies and estimators have not yet been adopted.
First, although the literature offers many different statistical techniques, CV practitioners do not have guidance towards a ``backpocket method'' that they can trust out-of-the-box.
Second, there is no comprehensive study that compares sampling and estimation strategies on CV data. 
Thus, it is not clear whether the additional complexity of using such sampling techniques will pay off in terms of lower costs.

We address both issues here.
We aim to give a readable and systematic account of methods from the statistics literature, test them on a large palette of CV models and datasets, and make a final recommendation for a simple and efficient method that the community can readily adopt. 
We take a practical point of view and choose to focus on one-shot selection techniques, rather than sequential sampling. 
This is because the job of annotating data is typically contracted out and carried out all at once, and thus the process of data sampling has to take place entirely before data annotation.

\begin{figure}[t]
    \centering
    \includegraphics[width=\textwidth]{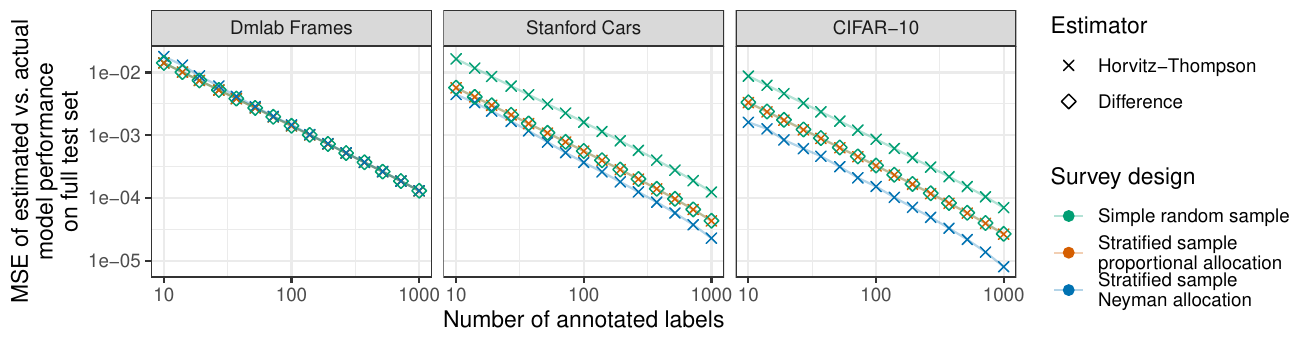}
    \caption{
        \textbf{Mean squared errors (MSEs) of estimators across sampling designs.}
        Estimates of zero-shot accuracy of ViT-B/32 in classification tasks on three datasets as a function of the amount of labeled data. 
        Stratified sampling can dramatically reduce the number of annotations needed to accurately estimate the model accuracy compared to the naive average ($\HT$) under simple random sampling. 
        Neyman allocation can sometimes further improve precision compared to proportional allocation.  
        (From left to right) No savings on the Dmlab Frames dataset, about 5x savings on the Stanford Cars, and about 10x savings on CIFAR-10.
        Note that the efficiency (precision) gains vary considerably between datasets (analysis and discussion in \Cref{sec:results}). 
        In the absence of stratified sampling with \kmeans on model predictions, the difference estimator can also greatly help. 
    } 
    \label{fig:intro_figure}
\end{figure}

More specifically, our work outlines a framework consisting of stratification, sampling design, and estimation components that practitioners can utilize when evaluating model performance.  
We review simple and stratified random sampling strategies with proportional and (optimal) Neyman allocation, as well as the Horvitz-Thompson and model-assisted difference estimators \cite{breidt2017model}.
Building on the survey sampling literature, we describe how to stratify the sample and design sampling strategies tailored to maximize the efficiency of the target estimator. 
We show that one should leverage accurate predictions of model performance (e.g., the predicted classification error of a CV classifier) in the stratification procedure or via the difference estimator to increase the precision of the estimates and reduce the number of samples needed for testing. 
We experimentally show how to apply the framework to benchmark models on CV classification tasks.

\Cref{fig:intro_figure} shows the main takeaways from our work. 
The model-assisted difference estimator and stratified sampling strategies (both proportional and Neyman) can significantly improve the precision of CV classifier accuracy estimates compared to naive averaging (Horvitz-Thompson) on a random data subset, e.g., achieving a 10x gain in precision on CIFAR-10 \cite{krizhevsky2009learning}. 
While the improvements may vary (e.g., modest gains on Stanford Cars \cite{KrauseStarkDengFei-Fei_3DRR2013}) or be less pronounced in some cases (e.g., DMlab Frames \cite{zhai2019visual}), stratified sampling with proportional allocation consistently offers a reliable and often superior performance.
Can one predict when clever methods will yield more bang for the buck? 
Yes, in \Cref{sec:results}, we explore the question of when and why these methods provide the most benefit to gain insights to apply them effectively.

\paragraph{Contributions and outline.\hspace{-0.5em}} 
A summary of our contributions and paper outline is as follows:
\begin{itemize}
[leftmargin=7mm]
    \item In \Cref{sec:setup}, we prescribe a statistical framework for model evaluation consisting of stratification, sampling, and estimation components. (\Cref{alg:metaalgo}).
    \item In \Cref{sec:methods}, we discuss the design of stratification, sampling, and estimator. 
    In particular, we show that maximizing the efficiency of the Horvitz-Thompson estimator under proportional allocation is equivalent to optimizing a \kmeans criterion (\Cref{prop:fin-min-to-pop-min,cor:k-means-optimal}).
    \item In \Cref{sec:results}, we explore the behavior of different options using a wide range of experiments on CV datasets. 
    We find that carefully designed stratification strategies as well as model-assisted estimators always yield more precise estimates of model performance compared to naive estimation under simple random sampling (\Cref{fig:in_distribution_results}). 
    Calibration and accurate prediction of the loss are key to obtaining highly efficient estimators (\Cref{fig:characterizing_efficiency}).
\end{itemize}

%% file: sections/related_work.tex
\section{Related Work}
\label{sec:related_work}

The idea of using clever sampling and estimators to obtain more precise estimates of a target of interest on a dataset has been extensively studied in the fields of survey sampling and machine learning. 
We review the relevant literature in these areas below. 

\subsection{Related Work in Survey Sampling}

The question of efficient or precise evaluation is essentially analogous to problems encountered in survey sampling \cite{fuller2011sampling, sarndal2003model, lohr2021sampling, cochran1977sampling, graubardand2002inference, isaki1982survey}
Survey sampling has two main inference paradigms. 
The first, design-based inference, views the dataset as static and assumes randomness only in sample selection. 
A well-known estimator in this framework, which we focus on, is the Horvitz-Thompson estimator \cite{horvitz1952generalization}, which averages the labeled samples reweighed by their propensity to be sampled. 
The second, model-based inference, assumes that the data are drawn from a superpopulation and uses statistical models for inference, leading to more precise estimates when the model is well-specified and less precise estimates when the model is misspecified.
The model-assisted approach combines the strengths of both by integrating modeling into the design-based framework. 
This approach yields (nearly) unbiased estimates as in the design-based paradigm but that are more precise when the model is correct. 
We focus on a popular model-assisted estimator, the difference estimator \cite{lumley2011connections, wu2001model, sarndal2003model, sarndal2007calibration}.
Our findings align with existing survey sampling literature \cite{breidt2017model, breidt2005model, mcconville2017model}, demonstrating that model-assisted estimators can significantly improve the precision of model performance estimates when predictions on the unlabeled sample are accurate.

Another crucial element in survey sampling is the design of the sample collection itself, which should aim to maximize the efficiency of the target estimator \cite{hajek1959optimal, chen2022optimal, chen2020optimal, imberg2023optimal, clark2022sample}.
There is a wide range of sampling designs (i.e., probability distributions over all possible samples), each designed to meet specific needs and contexts, together with the corresponding estimators \cite{tille2020sampling, brus2022spatial}. 
In this paper, we focus on simple random sampling with and without stratification because of its easily understandable advantages and trade-offs. 
We bypass more complex strategies such as unequal probability sampling, which can be carried out along with stratification, as they offer minimal additional benefits compared to stratified sampling with Neyman allocation when the number of strata is large \cite{neyman1992two}.

\subsection{Related Work in Machine Learning}

Efficient data sampling and estimation techniques have also been extensively discussed in the machine learning literature, particularly in the following settings.

\emph{Model performance estimation with fewer labels.} 
Multiple works have employed design-based estimators and considered the (active) setting where the labels are sampled iteratively \cite{li2019boosting, imberg2022active}. 
The devised sampling designs generally rely on stratification or unequal probability sampling, using predictions of model accuracy generated by the model itself \cite{sawade2010risk, sawade2010, Poms_2021_ICCV} or by a surrogate model \cite{kossen2021active, kossen2022active}. 
While our work shares many similarities with this line of research, we specifically focus on scenarios where labels are selected simultaneously, (mathematically and empirically) compare findings from different classes of estimators, offering practical advice on the best way to stratify. 
In addition, while these works focus on a few selected datasets, we compare the methods through a comprehensive array of experiments (see \Cref{sec:results}). 

\emph{Model performance estimation on unlabeled or partially labeled data.} 
Our paper is related to efforts on the estimation of model performance on unlabeled data \cite{chen2021mandoline, deng2021does, wenzel2022assaying, miller2021accuracy, Chen_2023_CVPR, yu2022predicting, chouldechova2022unsupervised}. 
These works focus on the prediction of classification accuracy on out-of-distribution data, leveraging indicators of distribution shift between training and test data such as Fr\'{e}chet distance \cite{deng2021labels}, discrepancies in model confidence scores between validation and test data \cite{guillory2021predicting, garg2022leveraging}, and disagreement between the predictions made by multiple similar models \cite{chuang2020estimating, jiang2021assessing, baek2022agreement}. 
A key takeaway from these works is that accurate estimation is a byproduct of proper model calibration \cite{wald2021calibration}, which is itself an area of active research \cite{ kim2022universal, roth2022uncertain, hebert2018multicalibration}. 
Some studies also address this challenge using a mix of unlabeled and labeled data, applying parametric models to predictions and existing labels \cite{Welinder_2013_CVPR, pmlr-v88-miller18a}. 
Notably, recent research has explored ``prediction-powered'' inference, a class of estimators that uses model predictions on the unlabeled data in the estimation process \cite{angelopoulos2023prediction, angelopoulos2023ppi++, zrnic2024active, zrnic2024cross}. 
In the case of mean estimation, this coincides with the model-assisted difference estimator from survey sampling. 
This line of work focuses on simple random sampling and Poisson sampling designs.
We contribute to this literature by comparing the performance of the difference estimator across stratified sampling methods. Our results show that, when stratified designs are used, the difference and Horvitz-Thompson estimators perform similarly. 

\emph{Active learning.} 
Our paper is also related to the literature on pool-based active learning, where the goal is to minimize the number of labels that are needed to ensure that the model achieves a given predictive accuracy \cite{settles2009active, cohn1996active, lewis1995sequential, chu2011unbiased, siddhant2018deep,ganti2012upal, farquhar2021statistical}. 
This is done by iterating between sampling and retraining. 
Traditionally, sampling designs in this area have focused on the predictive uncertainty of the model \cite{imberg2020optimal}, selecting instances one at a time \cite{lewis1994heterogeneous, scheffer2001active}. 
More recent work has explored other approaches \cite{gal2017deep, sener2017active, ren2021survey} and batch sampling strategies \cite{kirsch2019batchbald, ash2019deep}. 
Sampling strategies for model training and evaluation share many similarities. 
However, while (optimal) sampling designs tailored towards evaluation prioritize the sampling of data where model performance is most uncertain, active learning sampling approaches favor the sampling of observations that are anticipated to boost model performance. 

%% file: sections/framework.tex
\section{Framework Overview}
\label{sec:setup}

We provide a formal description of the problem setup and of our framework in \Cref{sec:formal_setup} and \Cref{sec:framework_overview} respectively. 
To ground our discussion, we use a classification task as a recurring example, although our framework also applies to regression tasks.

\subsection{Formal Setup}
\label{sec:formal_setup}

Consider a dataset $\cD$ consisting of $N$ instances $\{(X_i, Y_i) \colon i = 1, \dots, n\}$ drawn independently from distribution $P$. 
(Think of each instance as an image $X_i \in \cX$ and its corresponding ground truth label $Y_i \in \mathcal{Y}$.)
We have access to a predictive model $f$ that outputs estimates $\hatm_{y}(X_i)$ of the likelihood that label $y \in \mathcal{Y}$ is present in the $i$-th image $X_i$ for all $y \in \mathcal{Y}$. 
The predicted label with the highest score is $\hat{Y}_i = \argmax_{y \in \mathcal{Y}} \hatm_y(X_i)$.
Let $(X,Y)$ be a draw from $P$ and let $Z$ be the predictive error of our model $\hatm$ on $(X,Y)$.
Our target of interest is a predictive performance metric $\theta$ of the model $\hatm$, defined as $\theta = \mathbb{E}_P[Z]$. 
For example, taking $Z = \mathds{1}(Y = \hat{Y})$ yields the usual classification accuracy, $Z = (1 - \hatm_Y(X))^2$ the squared error, and $Z = - \log \hatm_Y(X)$ the cross-entropy.

In principle, we could estimate $\theta$ using $\cD$ by $\hat\theta_\cD = N^{-1}\sum_{i \in \mathcal{D}} Z_i$. 
However, while we have access to $X$ and to the outputs of $\hatm$ for all $1 \leq i \leq N$, $Y$ is not readily available. 
Our budget only allows us to obtain $Y$ for a subset of the instances $\cS \subset \cD$ of size $n \ll N$. 
We will randomly select these instances according to a sampling design $\pi$, which is a probability distribution over all subsets of size $n$ in $\cD$. 
We denote by $\pi_i > 0$ the likelihood that the $i$-th instance is included in $\cS$. 
Using the available data, we then obtain an estimate $\hat\theta$ of $\hat\theta_\cD$.

We measure the \emph{efficiency} of the estimator $\hat\theta$ of $\theta$ in terms of its mean squared error $\mse(\hat\theta, \theta) = \mathbb{E}_P[\mathbb{E}_\pi[(\hat\theta - \theta)^2]]$. 
The bias-variance decomposition yields $\smash{\mse(\hat\theta, \theta) \approx (\mathbb{E}_P[\mathbb{E}_\pi[\hat\theta]] - \hat\theta_\cD)^2 + \mathbb{E}_P[\Var_\pi(\hat\theta)]}$ because $\Var_P(\mathbb{E}_\pi[\hat\theta])$ is small when $n \ll N$. 
Thus, the $\mse$ will be driven by the bias and variance over the sampling design. 
Since we will only look at design-unbiased estimators, the $\mse$ will correspond to the variance. 
We define the relative efficiency of estimator $\hat{\theta}^{(1)}$ relative to $\hat{\theta}^{(2)}$ under a sampling design $\pi$ as the inverse of the ratio of their MSEs, that $\mse_\pi(\hat{\theta}^{(2)}, \hat{\theta}_\cD)/\mse_\pi(\hat{\theta}^{(1)}, \hat{\theta}_\cD)$. 
We say that estimator $\hat{\theta}^{(1)}$ is more efficient than $\hat{\theta}^{(2)}$ when the relative efficiency is greater than one.
 
\subsection{Framework Overview}
\label{sec:framework_overview}

\Cref{alg:metaalgo} outlines a framework for estimating the performance of a predictive model from a dataset $\cD$ when only a subset $\cS$ of instances has been labeled. 
The framework consists of an optional step for predicting model performance, a stratification or clustering procedure, a sampling design or strategy, and an estimator. 
Next, we discuss the choices for each of these components.

\begin{algorithm}[t]
\caption{A Framework for Efficient Model Evaluation (see \Cref{sec:framework_overview})}
\label{alg:metaalgo}
\begin{algorithmic}[1]
\REQUIRE Test dataset $\cD$ of size $N$ with  predictions of $\hatm$, annotation budget $n \ll N$.
\vspace{0.1em}
\STATE \textbf{Predict}: Construct a proxy $\hat{Z}$ of $\mathbb{E}_P[Z\,|\,X]$ and add predictions $\{\hat{Z}_i\}_{i\in \cD}$ to $\cD$.
\STATE \textbf{Stratify}: Partition the dataset into $H$ strata (or clusters) $\{\cD_h\}_{h=1}^H$ using $\hat{Z}$ or $X$.
\STATE \textbf{Sample}: Select $\cS$ ($|\cS|=n$) from $\cD$ based on the chosen design. 
\STATE \textbf{Annotate}: Obtain labels $\{Y_i\}_{i\in \cS}$, compute performance $\{Z_i\}_{i\in\cS}$. 
\STATE \textbf{Estimate}: Compute estimate $\hat\theta$ of model performance $\theta = \mathbb{E}_P[Z]$.
\smallskip
\ENSURE Estimate $\hat\theta$.
\smallskip
\end{algorithmic}
\end{algorithm}

\paragraph{Prediction (of $Z$).\hspace{-0.5em}} 
The first step involves building a proxy $\hat{Z}$ of $Z$ that is \emph{independent} of the observed labels. 
For example, when $Z = \mathds{1}(Y=\hat{Y})$ represents the accuracy of the classifier, we could take $\smash{\hat{Z}_i := \mathbb{E}[Z_i \, | \, X_i] = \mathbb{P}(Y_i = \hat{Y}_i)}$, which can be estimated via:
\begin{itemize}[leftmargin=7mm]
    \item \emph{Model predictions $\hatm(X)$}: 
    Use $\auxmod_i = \hatm_{\hat{Y}_i} (X_i)$ (likelihood of the model's top class) as a proxy. 
    \item \emph{Auxiliary predictions $\hatm^*(X)$}: 
    Use $\hat{Z}_i = \smash{\hatm^*_{\hat{Y}_i} (X_i)}$, the prediction of an auxiliary model $\hatm^*$ that, similarly to $\hatm$, estimates the probability distribution of 
    $Y$. 
\end{itemize}

\paragraph{Stratification.\hspace{-0.5em}}
Stratification involves partitioning the population $\cD$ into $H>0$ strata, $\{\cD_h\}_{h=1}^H$ with $|\cD_h|=N_h$. 
We can use standard clustering algorithms to form the strata based on: 
\begin{itemize}[leftmargin=7mm]
    \item \emph{Proxy $\hat{Z}$}: 
    Construct strata using the estimates $\{\hat{Z}_i\}_{i\in\cD}$.
    \item \emph{Features $X$}: 
    Cluster the images $\{X_i\}_{i\in \cD}$, e.g., by using their feature representations obtained from an encoder.
\end{itemize}

\paragraph{Sampling.\hspace{-0.5em}}
Two popular classes of sampling designs with fixed size and without replacement are: 
\begin{itemize}[leftmargin=7mm]
    \item \emph{Simple random sampling $(\SI)$}: Randomly sample $n$ instances from $\cD$ with equal probability.  
    \item \emph{Stratified simple random sampling $(\ST)$}: Allocate budget $n_h$ to each stratum $1\leq h \leq H$ such that $\sum_{h=1}^H n_h=n$ and conduct $\SI$ within each stratum, obtaining $\cS_h$. $\ST$ designs differ in how the budget $n$ is allocated to strata. We analyze two allocation strategies in \Cref{sec:methods}. Throughout our discussion, we will assume that strata sizes are large: $1/N_h\approx 0\; \forall h\in [H]$. 
\end{itemize}

\paragraph{Estimation.\hspace{-0.5em}}
We consider two instances of (unbiased) design-based and model-assisted estimators, chosen for their well-established statistical properties.\footnote{In preliminary experiments we assessed other model-assisted estimators in the class of ``generalized'' regression estimators \cite{sarndal2003model, wu2001model, breidt2017model} and found results comparable to $\DF$.}

\begin{itemize}[leftmargin=7mm]
    \item \emph{Horvitz-Thompson estimator $(\HT)$} \cite{horvitz1952generalization}: 
    This design-based estimator is defined as:
    \begin{equation} 
    \label{eq:estim_ht} 
    \hat\theta_{\HT} = \frac{1}{N}\sum_{i\in \cS} \frac{Z_i}{\pi_i}. 
    \end{equation} 
    $\HT$ is design-unbiased, that is, $\mathbb{E}_\pi[\hat\theta_{\HT}] =\hat\theta_\cD$ for $\pi\in\{\SI, \ST\}$. It follows that $\mse (\hat\theta, \theta) \approx  \mathbb{E}_P[\mse (\hat\theta, \hat\theta_\cD)] = \mathbb{E}_P[\Var_\pi(\hat\theta)]$ and the design-based variance is the sole source of error. 
    \item \emph{Difference estimator $(\DF)$} \cite{sarndal2003model, breidt2017model}:
    This model-assisted estimator is defined as:
    \begin{align}
    \hat\theta_{\text{DF}} = \frac{1}{N} \sum_{i \in \mathcal{D}} \auxmod_i + \frac{1}{N} \sum_{i \in \mathcal{S}} \frac{Z_i - \auxmod_i}{\pi_i},
    \end{align}
    where $\auxmod_i$ is an estimate of $Z_i$. The first term, $\sum_{i \in \mathcal{D}} \auxmod_i$, is independent of the sampling strategy. The second term corrects the bias of the first term, as $\sum_{i \in \mathcal{S}} \mathbb{E}_\pi [(Z_i - \auxmod_i)/\pi_i] = \sum_{i \in \mathcal{D}} (Z_i - \auxmod_i)$. This makes the $\DF$ estimator also unbiased under the sampling design.
\end{itemize}
These estimators offer complementary strengths: $\HT$ offers simplicity and unbiasedness, while $\DF$ provides potential variance reduction by incorporating model predictions. 
Our framework leverages these properties to improve the efficiency of model performance evaluation in computer vision tasks.
In the next section, we will discuss the optimal design of these stratification, sampling, and estimation components.

%% file: sections/design.tex
\section{Design of Framework Components}
\label{sec:methods}

To evaluate the effectiveness of the components in determining the estimator's variance or efficiency, we review the optimality of each.
In \Cref{sec:basic_efficiency} we analyze the efficiency of the estimators under the sampling designs. 
In \Cref{sec:stratification_theory} we discuss how the choice of the proxy $\hat{Z}$ for $Z$ can improve the efficiency of stratified sampling procedures. Lastly, in \Cref{sec:estimator_theory} we discuss the choice of the proxy in terms of the efficiency of the $\DF$ estimator. 

\subsection{Choosing the Sampling Design}
\label{sec:basic_efficiency}

Under $\SI$, $\pi_i = n/N$ for all $1\leq i \leq N$ and the $\HT$ estimator is simply the traditional empirical average $n^{-1}\sum_{i\in \cS}Z_i$. 
Its $\mse$ under the sample design is given by:
\begin{align}\label{eq:var_ht_srs}
 \textrm{MSE}_{\SI} (\hat\theta_{\HT}, \hat\theta_\cD) = \frac{1-f}{n}S_Z^2,
\end{align}
where $f=n/N$ represents sampling fraction and $S_Z^2= (N-1)^{-1}\sum_{i\in \cD} (Z_i - \hat\theta_\cD)^2$ is the variance of $Z$ in the finite population. 
From standard arguments in sampling statistics, under the setup of \Cref{sec:setup}, it can be shown that
\[
\mse_{\SI} (\hat\theta_{\HT}, \hat\theta_\cD)^{-1/2}(\hat\theta_{\HT} - \hat\theta_\cD) \overset{d}{\rightarrow} \cN(0, 1)
\]
as $n, N \rightarrow \infty$ and $N - n \rightarrow 0$ (see, e.g., Corollary 1.3.2.1 in \cite{fuller2011sampling}).
Estimation of the uncertainty around $\hat\theta_\HT$ can be performed using a plug-in estimator of the variance $S^2_Z$ in \eqref{eq:var_ht_srs}. 
In particular, when $f\approx 0$ and $1/n\approx 0$, we recover the common $\mse$ or variance estimator $\mse_\SI(\hat\theta_\HT, \hat\theta_\cD) \approx \sum_{i\in \cS}(Z_i - \hat\theta_\HT)^2/n^2$. 

One standard $\ST$ approach to budget splitting is proportional allocation, which assigns the budget proportionally to the size of the stratum in the finite population.
For all $1\leq h\leq H$, we assign $n_h \propto N_h$ and set $\pi_i = n_h / N_h$ for all $i \in \cD_h$. 
Under this allocation, the $\HT$ estimator is $\hat\theta_\HT = N^{-1}\sum_{h=1}^H(N_h/n_h)\sum_{i\in \cS_h} Z_i$ and its $\mse$ is given by: 
\begin{equation}
    \label{eq:var_ht_prop}
    \mse_{\ST, p} (\hat\theta_{\HT}, \hat\theta_\cD) = \frac{1-f}{n}\sum_{h=1}^H \frac{N_h}{N}S^2_{Z_h}, 
\end{equation}
where $S^2_{Z_h}$ is the variance of $Z$ in the $h$-th stratum \cite{tille2020sampling}. 
Analogous to $\SI$, asymptotic guarantees for $\HT$ under $\ST$ can also be obtained (see Theorem 1.3.2 in \cite{fuller2011sampling}). 

One can also seek a budget allocation that minimizes the error of $\HT$, which is $\mse(\hat\theta_\HT, \hat\theta_\cD)$. 
This strategy is known as Neyman or optimal allocation \cite{neyman1992two} and in the case of the $\HT$ estimator it assigns $\smash{n_h \propto N_h \sqrt{S^2_{Z_h}}}$ \cite{tille2020sampling, cochran1977sampling}. 
This means that more samples will be assigned to larger and more variable strata compared to proportional sampling. 
The $\HT$ estimator remains the same as under proportional allocation but its $\mse$ now becomes:
\begin{equation}\label{eq:var_ht_neyman}
    \mse_{\ST, o}(\hat\theta_{\HT}, \hat\theta_\cD) 
    = 
    \frac{1}{n} \bigg(\sum_{h=1}^N \frac{N_h}{N}S_{Z_h}\bigg)^2 - \frac{1}{N} \sum_{h=1}^N \frac{N_h}{N} S^2_{Z_h}.
\end{equation}
Since $\auxmod$ does not depend on the labels in $\cS$, the $\mse$s of the $\DF$ estimator under $\SI$ and $\ST$ are obtained by replacing $Z$ with $(Z-\hat{Z})$ in the formulas above, including for Neyman allocation \cite{chen2022optimal}. 

By comparing \eqref{eq:var_ht_srs}, \eqref{eq:var_ht_prop}, and \eqref{eq:var_ht_neyman}, we can derive the following well-known result, which identifies the sampling designs that yield the most precise estimates of $\hat\theta_\cD$ \cite{cochran1977sampling, tille2020sampling, lohr2021sampling, fuller2011sampling}.

\begin{proposition}
    \label{prop:mse_ordering}
    Under the setup of \Cref{sec:setup}, 
    \begin{equation}
        \label{eq:mse_ordering}
        \mathrm{MSE}_{\ST, o}(\hat\theta_{\HT}, \hat\theta_\cD) \leq  \mathrm{MSE}_{\ST, p} (\hat\theta_{\HT}, \hat\theta_\cD) \leq  \mathrm{MSE}_{\SI} (\hat\theta_{\HT},  \hat\theta_\cD). 
    \end{equation}
\end{proposition}

Similar inequalities also hold for the $\DF$ estimator. 
This result establishes that $\ST$ with proportional allocation consistently yields estimates with equal or lower $\mse$ compared to $\SI$ for the $\HT$ and $\DF$ estimators. 
The reduction in $\mse$ depends on the homogeneity of the strata:
When model performances $Z$ within each stratum are mostly equal, gains in efficiency of $\ST$ compared to $\SI$ are largest. 
When we know the standard deviation $S_{Z_h}$ and this term varies substantially across strata, Neyman allocation can provide even more precise estimates than proportional sampling. 
However, when our estimates of $S_{Z_h}$ are incorrect, Neyman allocation may lead to less precise even compared to $\SI$. 
The empirical results presented in \Cref{sec:results} align with these conclusions.

\subsection{Designing the Strata}\label{sec:stratification_theory} 

We turn to the construction of the strata. 
We can rewrite the $\mse$ of the $\HT$ estimator under $\ST$ in \eqref{eq:var_ht_prop} with proportional allocation as:
\small
\begin{equation}
    \label{eq:decomposition_msessrs}
    \mse_{\ST, p}(\hat\theta_{\HT}, \hat\theta_\cD)\approx 
    \frac{1-f}{Nn} \bigg\{\sum_{h=1}^H \sum_{i\in \cD_h} \big[(\hat{Z}_i - \hat{\bar{Z}}_{\cD_h})^2 
     + (\hat{\theta}_{\cD_h}^2- \hat{\bar{Z}}^2_{\cD_h}) \big]
     + \sum_{i\in \cD} \big[ (Z_i - \hat{Z}_i)^2
      + 2\hat{Z}_i(Z_i - \hat{Z}_i)\big] \bigg\},
\end{equation}
\normalsize
where $\hat{\bar{Z}}_{\cD_h} = N_h^{-1}\sum_{i\in\cD_h} \hat{Z}_i$ and $\hat\theta_{\cD_h} = N^{-1}_h \sum_{i\in\cD_h} Z_i$. 
The first term on the right-hand side of \eqref{eq:decomposition_msessrs} represents the within-strata sum of squares of the predictions $\{\hat{Z}_i\}_{i\in\cD}$. When $\hat{Z}_i \approx Z_i$, the second term becomes negligible. Since the remaining terms do not depend on the stratification, the strata construction affects the $\mse$ only through the first term. This intuition is formalized in the following result.

\begin{proposition}
    \label{prop:fin-min-to-pop-min}
    Assume that $\hat{Z}_i =\EE_P[Z_i\,|\,X_i]$ for all $i\in \cD$. Then the partition $\{\cD_h\}_{h=1}^H$ of $\cD$ that minimizes $\sum_{h=1}^H (N_h/N) S^2_{\hat{Z} h}$ also minimizes the error $\EE_P[\mse_{\ST, p}(\hat{\theta}_{\HT}, \hat{\theta}_\cD) \, | \, X]$ where $\smash{X=\{X_i\}_{i\in\cD}}$.
\end{proposition}

The result follows from standard decompositions for proper scoring rules \cite{kull2015novel} and implies that minimizing the weighted within-strata sum of squares for $\hat{Z}$ will also minimize the $\mse$ of the $\HT$ estimator under proportional stratification. 
In other words, this means that when a good predictor of the model performance based on $X$ is available, using its predictions \emph{alone} (as compared to clustering on $X$) will be sufficient to maximize the efficiency of the $\HT$ estimator.
This also provides practical guidance on which criterion to optimize, as summarized in the following corollary.
\begin{corollary}
    \label{cor:k-means-optimal}
    The partition $\{\cD_h\}_{h=1}^H$ of $\cD$ that minimizes $\EE_P\big[\mse_{\ST, p}(\hat{\theta}_{\HT}, \hat{\theta}_\cD)\,|\,X\big]$ is the same as that optimized by \kmeans clustering on $\{\EE_P[Z_i\,|\,X_i]\}_{i\in\cD}$. 
\end{corollary}

The corollary follows directly from \Cref{prop:fin-min-to-pop-min}. 
Thus, we can expect larger efficiency gains for the $\HT$ estimator under $\ST$ with proportional allocation when strata are formed by solving the \kmeans clustering criterion on $\EE_P[Z\,|\,X]$. 
In practice, this expectation is unknown and we have to rely on its proxy $\auxmod$.
A natural choice for the clustering algorithm is then to use the \kmeans algorithm itself on the proxy. 
Nonetheless, the experiments in \Cref{sec:methods} will show that even with estimated values, this approach still leads to better efficiency gains compared to stratifying based on the feature representations of $X$ obtained using the same model.

\subsection{Choosing the Estimator}
\label{sec:estimator_theory}

Based on our discussion in \Cref{sec:basic_efficiency}, one might have guessed that the $\DF$ estimator will have lower $\mse$ than the $\HT$ estimator when $Z$ and $\hat{Z}$ are positively associated.
To formally characterize this intuition, consider $\SI$, under which the $\mse$ of the $\DF$ estimator is:
\begin{align}
    \label{eq:antvar_si_df}
    \mse_{\SI} (\hat\theta_{\DF}, \hat\theta_\cD)
    \approx \frac{1-f}{n} \bigg\{ \frac{1}{N}\sum_{i\in \cD}(Z_i - \hat{Z}_i)^2 - (\hat\theta_\cD-\hat{\bar{Z}})^2\bigg\}.
\end{align}
The first term on the right-hand side of \eqref{eq:antvar_si_df} represents the $\mse$ of $\hat{Z}_i$ with respect to $Z_i$, while the second term represents a squared calibration error. 
It follows that choosing $\hat{Z}_i = \EE_P[Z_i\,|\,X_i]$ for all $1\leq i \leq N$ minimizes the expected $\mse$ of $\DF$ under $\SI$.
This choice for $\hat{Z}$ aligns with our recommendation from the stratification procedure and leads to the following result: 
\begin{proposition}
    \label{prop:efficiency_df_vs_ht}
    Assuming $\hat{Z}_i=\mathbb{E}_P[Z_i\,|\,X_i]$, we have
    \begin{align}\label{eq:ratio_expected_mse_df_ht}
     \frac{\mathbb{E}_P[\mse_{\SI} (\hat\theta_{\DF}, \hat\theta_\cD)]}{\mathbb{E}_P[\mse_{\SI} (\hat\theta_{\HT}, \hat\theta_\cD)]} = \frac{\mathbb{E}_P[\Var_P(Z\,|\,X)]}{\Var_P(Z)}.
\end{align}
\end{proposition}
Since $\Var_P(Z)=\mathbb{E}_P[\Var_P(Z\,|\,X)] + \Var_P(\mathbb{E}_P[Z\,|\,X])$ by the law of total variance, the ratio in \Cref{prop:efficiency_df_vs_ht} will always be less than $1$. 
This means that the $\DF$ estimator will yield more precise estimates than $\HT$ as long as $\hat{Z}$ is well specified. 
The efficiency gains of $\DF$ over $\HT$ under $\SI$ will be highest  when the auxiliary information $X$ is predictive of $Z$, that is, when $\Var_P (\mathbb{E}_P [Z\,|\,X])$ is large.

Under $\ST$ with proportional allocation, we can similarly show that 
\[
\mse_{\ST, p}(\hat\theta_{\DF}, \hat\theta_{\cD})~\approx~\frac{1-f}{n}~\left\{\frac{1}{N}\\ \sum_{i\in\cD} (Z_i - \hat{Z}_i)^2 - \sum_{h=1}^H \frac{N_h}{N} (\hat\theta_{\cD_h} - \hat{\bar{Z}}_{\cD_h})^2\right\}.
\]
Since the first term on the right-hand side will in general dominate when the proxy is calibrated, we should not expect significant efficiency gains of $\DF$ compared to $\HT$ under this sampling design when the strata are finegrained enough, i.e., $\auxmod_i - \hat{\bar{Z}}_{\cD_h}\approx 0$ for all $i\in\cD_h$. 
Thus, the uncertainty of the $\DF$ and $\HT$ estimates under $\ST$ will be close, i.e.,  $\mse_{\ST, p} (\hat\theta_{\DF}, \hat\theta_\cD)\approx \mse_{\ST,p} (\hat\theta_{\HT}, \hat\theta_\cD)$.

%% file: sections/results.tex
\section{Empirical Evaluation}
\label{sec:results}

\subsection{Experimental Setup}

To evaluate the methods, we consider the classification setup described in \Cref{sec:setup}. 
Our goal is to compare the efficiency or precision of sampling designs and associated estimators of the predictive performance of a model $f$, namely $\theta=\EE_P[Z]$, by having access only to a limited number of labels (say $n=100$) from our test dataset $\cD$ of size $N\gg n$. 

\paragraph{Tasks and models.\hspace{-0.5em}}
Our main evaluation focuses on the zero-shot classification accuracy ($Z = \mathds{1}(Y = \hat{Y})$) of a CLIP model $f$ with ViT-B/32 as the visual encoder, pretrained on the English subset of LAION-2B \cite{ilharco_gabriel_2021_5143773, schuhmann2022laionb, radford2021learning}. 
We evaluate its accuracy on the tasks included in the LAION CLIP-Benchmark \cite{laion_clip_benchmark}; the full list is provided in \Cref{app:additional_results}. 
This benchmark covers a wide range of model performances and task diversities, making it a suitable testbed for comparing different estimation methods.
To construct $\hat{Z}$, we use the confidence scores from either CLIP ViT-B/32 or from the surrogate model $f^*$ CLIP ViT-L/14. 
The latter model achieves higher classification accuracy than the former across most tasks in the benchmark, meaning that the proxy $\hat{Z}$ is a better predictor of $Z$. 
Additionally, we calibrate the proxy $\hat{Z}$ with respect to $Z$ via isotonic regression on a randomly sampled half subset of $\cD$; technically, one could conduct training and evaluation on the same dataset with cross-fitting. 
We carry out the estimation procedure on the remaining half of the data, $\cD$.
For stratification purposes, we obtain feature representations from the penultimate layer of $f$. 
We arbitrarily set the number of strata to $10$ for all experiments; in the case of $\ST$ with proportional allocation, more strata would lead to more efficiency gains. 
Our code and package is available at \href{https://github.com/amazon-science/ssepy}{github.com/amazon-science/ssepy}.

\paragraph{Additional experiments.\hspace{-0.5em}}
In \Cref{app:additional_results}, we include additional experiments to evaluate the performance of our methods. These experiments cover: 
(\cref{app:experiments_othermetrics}) the estimation of performance metrics other than classification accuracy; 
(\cref{sec:experiments_lp}) results with predictions generated with linear probing and 
(\cref{app:experiment_otherarchitectures}) with predictions by CLIP with ResNet and ConvNeXT backbones \cite{he2016deep, liu2022convnet}; 
(\cref{app:experiment_ood}) an analysis 
on two datasets from the WILDS out-of-distribution benchmark \cite{koh2021wilds}. 
Some of the results of these experiments are also summarized in \Cref{sec:discussion}. 
In particular, we compare the efficiency of our methods on data that is out-of-distribution for the model and for the proxy $\hat{Z}$ of $Z$.

\subsection{Results}

We study the efficiency of sampling design, stratification procedures, and estimators. We then analyze where the efficiency gains over $\HT$ under $\SI$ arise. 

\begin{figure*}[t]
    \centering
    \includegraphics[width=\textwidth]{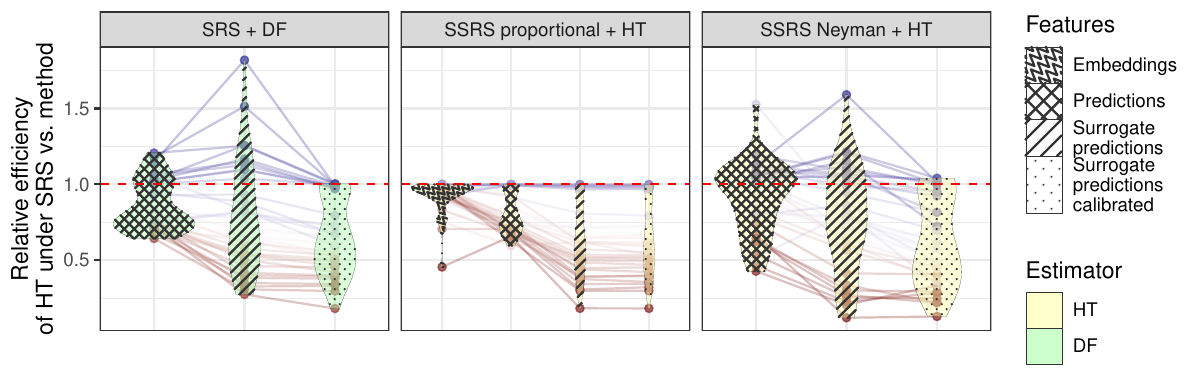}
    \caption{
        \textbf{Comparison of efficiency across stratification procedures, sampling designs, and estimators.} 
        The violin plots illustrate the relative efficiency of the Horvitz-Thompson ($\HT$) estimator under simple random sampling ($\SI$, red dashed line) compared to other survey sampling strategies and estimators (relative efficiency is $\textrm{MSE}_\pi(\hat\theta_{\text{\tiny{EST}}})/\textrm{MSE}_{\SI}(\hat\theta_{\HT})$) for estimating the accuracy of CLIP ViT-B/32 on classification tasks in the benchmark. 
        Lower values indicate larger efficiency gains compared to the baseline. 
        The dots and lines represent the relative efficiencies of the sampling methods and estimators on the various tasks.
    } 
    \label{fig:in_distribution_results}
\end{figure*}

\paragraph{Sampling design.\hspace{-0.5em}}
\Cref{prop:mse_ordering} states that estimates obtained through $\ST$ with proportional allocation using the $\HT$ and $\DF$ estimators consistently achieve lower variance or $\mse$ compared to those obtained via $\SI$. 
\Cref{fig:in_distribution_results} corroborates this analytical finding (see also the results in Table in \Cref{app:additional_results}), showing that $\mse_\ST(\hat\theta_\HT, \hat\theta_\cD)\leq \mse_\SI(\hat\theta_\HT, \hat\theta_\cD)$ regardless of the features used for stratification. 
The gain varies across tasks and, when using surrogate model predictions, the relative efficiency ranges from about 10x on some tasks to no gain on others. 
While Neyman allocation is guaranteed to yield more precise estimates compared to these sampling designs when the allocation is based on $S_{Z_h}$, in practice we need to rely on its plug-in estimator $\smash{\hat{S}_{Z_h} = [\hat{\bar{Z}}_{\cD_h} (1 - \hat{\bar{Z}}_{\cD_h})]^{1/2}}$. 
This can introduce inaccuracies in the budget allocation. 
Indeed, we observe that Neyman allocation can perform even worse than $\SI$ and $\ST$ with proportional allocation. 
However, when $\hat{Z}$ is derived from the predictions of the surrogate model $f^*$ and is further calibrated, then Neyman allocation consistently matches or exceeds the performance under $\SI$. 
On certain tasks, the $\mse$ of $\HT$ is more than 10x lower compared to under $\SI$.

\paragraph{Stratification.\hspace{-0.5em}}
In \Cref{sec:stratification_theory}, we discussed how stratifying on $\hat{Z}=\hatm_{\hat{Y}}(X)$ can result in higher homogeneity within strata compared to stratifying directly on the image embeddings obtained from the same model.
This is consistent with the findings in \Cref{fig:in_distribution_results}, where the efficiency of the $\HT$ estimator under $\ST$ with proportional allocation is generally higher when stratification is performed on the proxy. 
Stratification using the proxy based on the predictions generated by a surrogate model $f^*$ with higher performance, here CLIP ViT-L/14, additionally increases efficiency.
This improvement is observed for proportional allocation across all tasks and, in most cases, for Neyman allocation as well. 
Calibrating these predictions does not appear to affect the formation of the strata and therefore does not affect performance under proportional allocation. 
However, it does change the allocation of the budget and consequently, we observed an increase in the performance of the estimates under Neyman allocation. 

\paragraph{Estimator.\hspace{-0.5em}}
The analysis in \Cref{sec:stratification_theory} suggests that, under $\SI$, the $\DF$ estimator has the potential to significantly improve the precision of our estimates compared to $\HT$. 
However, as shown in \Cref{fig:in_distribution_results}, the efficiency gains of the $\DF$ estimator should not be taken for granted. 
When $\hat{Z}$ is based on uncalibrated model predictions, we observe that $\DF$ achieves higher efficiency than $\HT$ in many but not all of the tasks. In some cases, it performs substantially worse than $\HT$. 
However, the $\DF$ estimator that leverages the calibrated proxy always achieves equal or lower $\mse$ than HT. 
Consistently with our theoretical findings in \Cref{sec:estimator_theory}, the values of $\mse_{\SI} (\hat\theta_{\DF}, \hat\theta_\cD)$ for calibrated predictions are close to those of $\mse_{\ST,p} (\hat\theta_{\HT}, \hat\theta_\cD)$, indicating similar gains in efficiency. 
Finally, as discussed in \Cref{sec:estimator_theory}, $\HT$ and $\DF$ under $\ST$ yield estimates with virtually the same precision and are excluded from the figure.

\begin{figure*}[t]
    \includegraphics[width=0.49\textwidth]{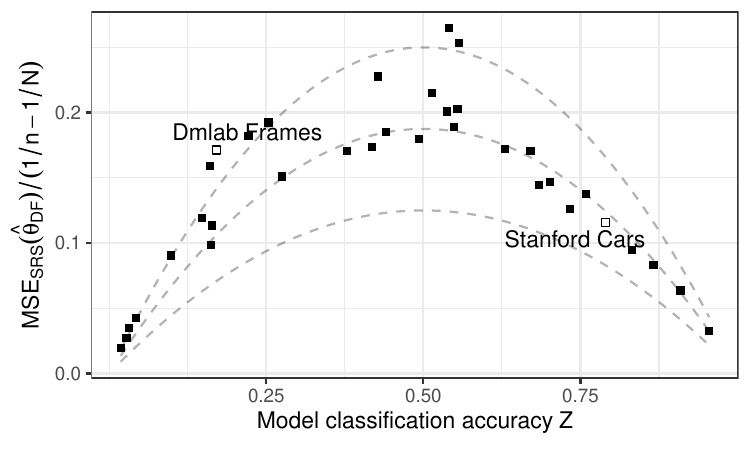}
    \includegraphics[width=0.49\textwidth]{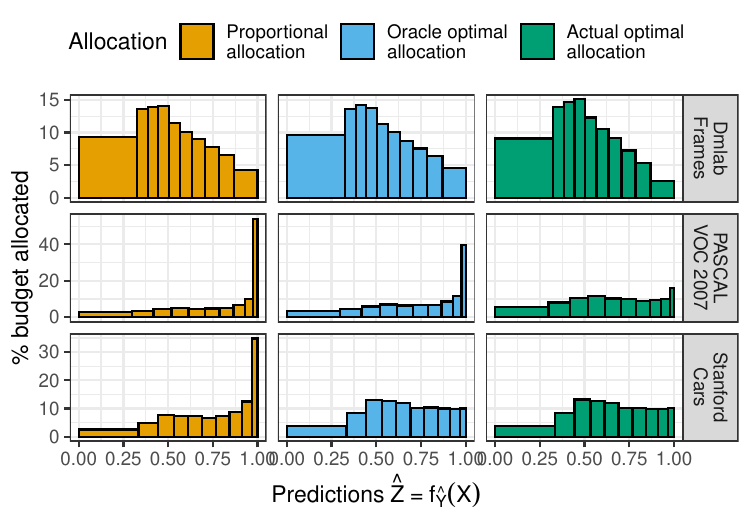}
    \caption{
    \textbf{Characterization of efficiency gains.}
    The left panel shows the mean squared error ($\mse$) of the difference estimator ($\DF$) under simple random sampling ($\SI$, corrected by $n/(1-f)$) as a function of the zero-shot classification accuracy $N^{-1}\sum_{i\in \cD}Z$ of CLIP ViT-B/32 evaluated on the full test sets of the LAION CLIP benchmark tasks. We construct $\hat{Z}$ using the predictions of CLIP with ViT-B/32 as backbones. 
    Dashed lines correspond to the relative efficiencies of $1$ (highest line), $0.75$, and $0.5$ (lowest). 
    In tasks where the model achieves higher classification accuracy, it also tends to have higher relative efficiency. 
    The right panel shows the allocation of the annotation budget to each stratum through proportional and optimal (ideal based on $S_{Z_h}$ and actual based on $\hat{S}_{\hat{Z}h}$) allocations across three datasets. 
    In practice, Neyman allocation provides efficiency gains over proportional allocation only on Stanford Cars.
    }
    \label{fig:characterizing_efficiency}
\end{figure*}

\paragraph{Characterizing the efficiency gains.\hspace{-0.5em}}
The empirical results presented so far indicate that the efficiency gains of the $\DF$ estimator and the stratified designs over the naive average of a completely random subset of data (i.e., $\HT$ under $\SI$) vary significantly across tasks. 
To determine when we can expect the largest gains, we turn to our theoretical analysis. 
In \Cref{sec:estimator_theory}, we have shown that larger efficiency gains for the $\DF$ estimator should be expected when the $\mse$ of $\hat{Z}$ relative to $\Var_{\SI}(\hat\theta_{\HT})\propto \hat\theta_{\cD}(1-\hat\theta_{\cD})$ is low or similarly when $\mathbb{E}_P[\Var_P(Z\,|\,X)]\ll \Var_P(Z\,|\,X)$ in \Cref{prop:efficiency_df_vs_ht}. 
Note that classifiers with the same $\Var_{\SI}(\hat\theta_{\HT})$ can have very different accuracy (e.g., $\hat\theta_\cD = 0.2$ vs. $\hat\theta_\cD = 0.8$) and classifiers with higher accuracy often achieve lower MSE, which will be associated with larger efficiency gains of $\DF$ over $\HT$ under $\SI$. 
This observation is confirmed by \Cref{fig:characterizing_efficiency}, where we observe that $\Var_{\SI}(\hat\theta_{\HT})$ is similar on Dmlab Frames and Stanford Cars, but $f$ achieves higher accuracy on the latter and also $\HT$ under $\SI$ yields more precise estimates of $\hat\theta_\cD$. 
It is worth noting that this argument may not always hold, as a classifier's high accuracy may be explained by extreme class imbalance. 
Nevertheless, in the tasks we have examined, this observation generally holds. 
Efficiency gains of $\DF$ over $\HT$ under $\SI$ are inherently tied to those of $\ST$ with proportional allocation, so similar arguments hold for that sampling design. 
We also mentioned in \Cref{sec:basic_efficiency} that Neyman allocation may not yield sizable gains over proportional if the $S_{Z_h}$'s (i) are similar across strata or (ii) are poorly estimated. 
\Cref{fig:characterizing_efficiency} shows two examples of (i) and (ii), as well as an example where Neyman allocation leads to large gains. 
On Dmlab Frames, (i) occurs: The distributions of $\hat{Z}$ conditional on $Z=0,1$ mostly overlap, hence proportional and Neyman allocation are similar. 
On Pascal VOC 2007, we observe (ii): Neyman allocates too little budget to the stratum where $\hat{Z}$ is close to $1$, which has considerable variability. 
Lastly, on Stanford Cars, Neyman allocation leads to a large gain, as the proportional allocation allocates too much budget to high values of $\hat{Z}$, even though the model makes few errors in that region (i.e., mostly $Z=1$).

%% file: sections/discussion.tex
\section{Discussion}
\label{sec:discussion}

In this paper, we have investigated methods to evaluate the predictive performance of a machine learning model on large datasets on which only a limited amount of data can be labeled.
Our findings show that, when good predictions of the model's performance are available, stratified sampling strategies and model-assisted estimators can provide more precise estimates compared to the traditional approach of naive averaging on a data subset obtained via $\SI$. 

\paragraph{Main takeaway.\hspace{-0.5em}}
We recommend that, when selecting a data subset to annotate, CV practitioners always use stratified sampling strategies ($\ST$) with proportional allocation, running \kmeans on the proxy $\hat{Z}$ of model performance $Z$ (\Cref{sec:stratification_theory}). 
The more strata one can form, the higher the precision of the estimates will likely be. 
When the proxy $\hat{Z}$ is well calibrated, Neyman allocation may also be used and may lead to additional efficiency gains (\Cref{sec:basic_efficiency}). 
If a data subset has already been obtained via $\SI$, then one can still leverage the $\DF$ estimator to increase the precision of the estimates (\Cref{sec:estimator_theory}). 
If there is uncertainty about the quality of the proxy, the method recently proposed by \cite{angelopoulos2023ppi++} can be applied to adjust the extent to which the estimator relies on the proxy.

Beyond that, it is important to understand the efficiency of these estimators on out-of-distribution data. 
In this setting, the proxy $\auxmod$ may be a poor predictor of model performance $Z$ and consequently, the gains of $\ST$ with proportional allocation or $\DF$ under $\SI$ relative to $\HT$ under $\SI$ may be limited.  
There is also the risk that $\ST$ with Neyman allocation may yield estimates that have substantially higher variance than those obtained under proportional allocation. 
Therefore, caution should be exercised when using adaptive allocations and one believes that the test distribution may differ from the training distribution.  
These findings suggest that incorporating calibration techniques (of models and estimators) \cite{sarndal2007calibration, yu2022robust} along with sequential sampling \cite{zrnic2024active} may lead to additional improvements in the evaluation of model performance.

\section*{Acknowledgments}

We thank the anonymous reviewers for their encouraging comments and valuable suggestions that have improved our manuscript. We also thank Tijana Zrnic for highlighting the connections between prediction-powered inference and our work, as well as Georgy Noarov for pointing out the link between our results and decompositions for proper scoring rules.

%% file: sections/appendix.tex
\begin{center}
\Large
{\bf \framebox{Appendix}}
\end{center}

\bigskip

This appendix complements our main paper ``A Framework for Efficient Model Evaluation through Stratification, Sampling, and Estimation.''

\section*{Organization}

The appendix is organized as follows.

\begin{itemize}[leftmargin=7mm]
    \item
    In \Cref{app:proofs}, we provide proofs of the theoretical results presented in the paper. Specifically, this section includes the following proofs:
    \begin{itemize}[leftmargin=7mm]
        \item Proof of \Cref{prop:mse_ordering} (\Cref{proof:prop:strata_equivalence_predictor-1}).
        \item Proof of \Cref{prop:fin-min-to-pop-min} (\Cref{proof:prop:fin-min-to-pop-min}).
        \item Proof of \Cref{prop:efficiency_df_vs_ht} (\Cref{proof:prop:efficiency_df_vs_ht}).
    \end{itemize}
    \smallskip
    \item
    In \Cref{app:additional_results}, we present additional results that complement the findings in the paper. 
    Specifically, this section includes the following results:
    \begin{itemize}[leftmargin=7mm]
        \item Breakdown of the results shown in \Cref{fig:in_distribution_results} (\Cref{app:experiments_breakdownresults}).
        \item Comparison of different methods for estimating classifiers mean squared error (MSE) and cross-entropy (\Cref{app:experiments_othermetrics}).
        \item Assessment of classification accuracy in CLIP models using linear probing (\Cref{sec:experiments_lp}).
        \item Tests on CLIP models with ResNet and ConvNeXT visual encoders (\Cref{app:experiment_otherarchitectures}).
        \item Further information on the out-of-distribution results presented in \Cref{fig:indvsood} (\Cref{app:experiment_ood}).
    \end{itemize}
\end{itemize}

\input{sections/proofs}

\clearpage

\input{sections/extended_results}

%% file: sections/proofs.tex
\section{Proofs}
\label{app:proofs}

\subsection{Proof of \Cref{prop:mse_ordering}}
\label{proof:prop:strata_equivalence_predictor-1}

This proof is standard and can be found in survey sampling textbooks \cite{cochran1977sampling, tille2020sampling}. 
For the reader's convenience, we provide the proof below in our notation.

\paragraph{Part 1.\hspace{-0.5em}}
To prove that 
$
    \mse_{\ST, p} (\hat\theta_\HT, \hat\theta_\cD)\leq \mse_{\SI} (\hat\theta_\HT, \hat\theta_\cD),
$
recall that $\mse_{\ST} (\hat\theta_\HT, \hat\theta_\cD) = \Var_{\ST, p} (\hat\theta_\HT)$ and $\mse_{\SI} (\hat\theta_\HT, \hat\theta_\cD) = \Var_{\SI} (\hat\theta_\HT)$ as the estimators are unbiased with respect to the sampling design. 
Thus, we need to show that 
$
    \Var_{\SI} (\hat\theta_\HT) - \Var_{\ST, p}(\hat\theta_\HT) \geq 0. 
$
We can rewrite
\[
    (N-1)S_Z^2 = \sum_{h=1}^H (N_h-1) S^2_{Z_h} + \sum_{h=1}^H N_h (\hat\theta_{\cD_h} - \hat\theta_{\cD})^2,
\]
where $\cD_h=N_h^{-1}\sum_{i\in\cD_h} Z_i$. 
When $(N_h-1)/N\approx N_h/N$, 
\[
   S^2_{Z_h} \approx  \sum_{h=1}^H \frac{N_h}{N} S^2_{Z_h} + \sum_{h=1}^{H} \frac{N_h}{N}  (\hat\theta_{\cD_h} - \hat\theta_{\cD})^2.
\]
Consequently, we have 
\[
    \Var_{\SI} (\hat\theta_\HT) - \Var_{\ST, p}(\hat\theta_\HT) \approx  \frac{1-f}{n} \sum_{h=1}^H \frac{N_h}{N}(\hat\theta_{\cD_h} - \hat\theta_{\cD})^2,
\]
completing the first part of the proof.

\paragraph{Part 2.\hspace{-0.5em}}
To show that 
$
    \mse_{\ST, o} (\hat\theta_\HT, \hat\theta_\cD)\leq \mse_{\ST, p} (\hat\theta_\HT, \hat\theta_\cD),
$
and equivalently
$
    \Var_{\ST, o} (\hat\theta_\HT)\leq \Var_{\ST, p} (\hat\theta_\HT),
$
we observe that 
\begin{align*}
    \Var_{\ST, p} (\hat\theta_\HT) - \Var_{\ST, o} (\hat\theta_\HT) 
    &= \frac{1}{n} \sum_{h=1}^H \frac{N_h}{N}S^2_{Z_h} - \frac{1}{n} \bigg(\sum_{h=1}^H \frac{N_h}{N}S_{Z_h} \bigg)^2 \\
    & =\frac{1}{n} \sum_{h=1}^H \frac{N_h}{N}(S_{Z_h} - \bar{S_{Z}})^2,
\end{align*}
where $\bar{S_{Z}}=\sum_{h=1}^H (N_h/N) S_{Z_h}$. 
This completes the second part of the proof. 

\subsection{Proof of \Cref{prop:fin-min-to-pop-min}}
\label{proof:prop:fin-min-to-pop-min}

Note: Analogous result can be also be derived (in more generality) by using the three-term decomposition of proper scoring rules in \cite[Section 5]{kull2015novel}.
Below we provide a proof using our notation for the setting of this paper.

Recall that the expected variance conditional on $X=(X_1, \dots, X_N)$ of the $\HT$ estimator under $\ST$ with proportional allocation is 
\begin{equation*}
    \EE_P[\Var_{\ST, p}(\hat\theta_\HT)\,|\,X] = \frac{1-f}{n}\sum_{h=1}^H \frac{N_h}{N} \frac{1}{N_h-1}\sum_{i\in \cD_h} \EE[ ( Z_i - \hat\theta_{\cD_h} )^2 \,|\, X ].
\end{equation*}
Now, let $\hat{Z}=\EE_P[Z\,|\,X]$ and $\hat{\bar{Z}}_{\cD_h} = N_h^{-1}\sum_{i\in\cD_h}\hat{Z}_i$. 
We can decompose $(Z_i - \hat\theta_{\cD_h})^2$ as follows:
\begin{align*}
    (Z_i - \hat\theta_{\cD_h})^2
    &= (Z_i - \hat{Z}_i + \hat{Z}_i -  \hat\theta_{\cD_h})^2 \\
    &= (Z_i - \hat{Z}_i)^2 + (\hat{Z}_i -  \hat\theta_{\cD_h})^2 + 2(Z_i - \hat{Z}_i)(\hat{Z}_i -  \hat\theta_{\cD_h})\\
    &= (Z_i - \hat{Z}_i)^2 + (\hat{Z}_i - \hat{\bar{Z}}_{\cD_h} + \hat{\bar{Z}}_{\cD_h} -  \hat\theta_{\cD_h})^2 + 2(Z_i - \hat{Z}_i)(\hat{Z}_i -  \hat\theta_{\cD_h})\\
    &= (Z_i - \hat{Z}_i)^2 + (\hat{Z}_i - \hat{\bar{Z}}_{\cD_h})^2 + (\hat{\bar{Z}}_{\cD_h} -  \hat\theta_{\cD_h})^2 \\ 
    & \quad + 2(\hat{Z}_i - \hat{\bar{Z}}_{\cD_h})(\hat{\bar{Z}}_{\cD_h}- \hat\theta_{\cD_h}) + 2(Z_i - \hat{Z}_i)(\hat{Z}_i -  \hat\theta_{\cD_h}).
\end{align*}

We can show that 
\begin{align*}
    \sum_{i\in\cD_h} (\hat{Z}_i - \hat{\bar{Z}}_{\cD_h})(\hat{\bar{Z}}_{\cD_h}- \hat\theta_{\cD_h})
    &= (\hat{\bar{Z}}_{\cD_h} - \hat\theta_{\cD_h}) \sum_{i\in\cD_h}(\hat{Z}_i - \hat{\bar{Z}}_{\cD_h}) = 0. 
\end{align*}
Then, since 
\[
    \frac{1}{N_h - 1} \sum_{i\in\cD_h}[ (\hat{\theta}_{\cD_h} - \hat{\bar{Z}}_{\cD_h})^2 + 2(Z_i - \hat{Z}_i)(\hat{Z}_i -  \hat\theta_{\cD_h})] \approx (\hat{\theta}_{\cD_h}^2 - \hat{\bar{Z}}^2_{\cD_h}) + 2\hat{Z}_i(Z_i - \hat{Z}_i),
\]
when $1/N_h\approx 0$, we obtain \eqref{eq:decomposition_msessrs}. 
However, we will continue the proof without this assumption. 

By the assumption of independence, we also have 
\begin{align*}
    &\EE_P[(Z_i - \hat{Z}_i)(\hat{Z}_i -  \hat\theta_{\cD_h})\,|\,X]\\
    &= - \EE_P[(Z_i - \EE_P[Z_i \, | \, X_i])(\hat\theta_{\cD_h} - \EE_P[Z_i \, | \, X_i])\,|\,X] \\
    &= - \frac{1}{N_h} \EE_P[(Z_i - \EE_P[Z_i \, | \, X_i])(Z_i - \EE_P[Z_i \, | \, X_i])\,|\,X] \\
    &= - \frac{1}{N_h}\Var_P(Z_i\,|\,X_i).
\end{align*} 
Using similar arguments, we obtain 
\begin{align*}
    \EE_P[(\hat{\bar{Z}}_{\cD_h} - \hat\theta_{\cD_h})^2\,|\,X] = \frac{1}{N_h^2} \sum_{i\in\cD_h} \Var_P(Z_i\,|\,X_i). 
\end{align*}
Thus, we have 
\begin{align*}
    &\sum_{i\in\cD_h} \EE_P[(Z_i - \hat\theta_{\cD_h})^2\,|\,X] \\
    &= \sum_{i\in\cD_h}\mbbE_P[(Z_i - \hat{Z}_i)^2\,|\,X] + \sum_{i\in\cD_h} (\hat{Z}_i - \hat{\bar{Z}}_{\cD_h})^2 - \frac{1}{N_h}\sum_{i\in\cD_h} \Var_P(Z_i\,|\,X_i)\\
    &= \sum_{i\in\cD_h}\Var_P(Z_i \, | \, X_i) + \sum_{i\in\cD_h} (\hat{Z}_i - \hat{\bar{Z}}_{\cD_h})^2 - \frac{1}{N_h}\sum_{i\in\cD_h} \Var_P(Z_i\,|\,X_i).
\end{align*}
Finally, the above implies that
\begin{align*}
     &\EE_P[\Var_{\ST, p}(\hat\theta_\HT)\,|\,X] \\
     &= \frac{1-f}{n}\sum_{h=1}^H \frac{N_h}{N(N_h-1)}\sum_{i\in \cD_h} \bigg\{\Var_P(Z_i\,|\,X_i) + (\hat{Z}_i - \hat{\bar{Z}}_{\cD_h})^2 - \frac{1}{N_h} \Var_P(Z_i\,|\,X_i)\bigg\}\\
     &=\frac{1-f}{n}\sum_{h=1}^H \frac{N_h}{N(N_h-1)}\sum_{i\in \cD_h} \bigg\{\frac{N_h-1}{N_h}\Var_P(Z_i\,|\,X_i) + (\hat{Z}_i - \hat{\bar{Z}}_{\cD_h})^2\bigg\}\\
     &= \frac{1-f}{n}\frac{1}{N}\bigg\{\sum_{i\in \cD} \Var_P(Z_i\,|\,X_i) + \sum_{h=1}^H \frac{N_h}{N_h-1}\sum_{i\in \cD_h}(\hat{Z}_i - \hat{\bar{Z}}_{\cD_h})^2 \bigg\}\\
     &=\frac{1-f}{n}\frac{1}{N}\sum_{i\in \cD} \Var_P(Z_i\,|\,X_i) + \frac{1-f}{n}\sum_{h=1}^H \frac{N_h}{N} S^2_{\hat{Z}h}.
\end{align*}
Note that the first term does not depend on the specific strata, hence the stratification procedure only affects the second term. This completes the proof.  

\subsection{Proof of \Cref{prop:efficiency_df_vs_ht}}
\label{proof:prop:efficiency_df_vs_ht}

We start with the decomposition in \eqref{eq:antvar_si_df}:
\begin{equation}
    \Var_\SI (\hat\theta_\DF) = \frac{1-f}{n} \bigg\{ \frac{1}{N-1} \sum_{i=1}^N (Z_i - \hat{Z}_i)^2 - \frac{N}{N-1}(\hat\theta_\cD - \hat{\bar{Z}})^2 \bigg\}.
\end{equation}
By the independence of $\hat{Z}_i$ and $Z_i$, we have
\begin{align*}
      N^2\mathbb{E}_P \big[(\hat\theta_\cD - \hat{\bar{Z}})^2 \,\big|\, X \big]
      &= \sum_{i=1}^N \mathbb{E}_P \big[ (Z_i - \hat{Z}_i)^2 \,|\, X \big] + \sum_{i=1}^N \sum_{\substack{j=1\\j\neq i}}^N \mathbb{E}_P\big[Z_i- \hat{Z}_i \,|\, X \big] \cdot \mathbb{E}_P\big[ Z_j - \hat{Z}_j \,|\, X \big]\\  
      &= \sum_{i=1}^N \mathbb{E}_P \big[ (Z_i - \hat{Z}_i)^2 \,|\, X \big]. 
\end{align*}
Therefore, we obtain
\begin{align*}
    \mathbb{E}_P\big[ \Var_\SI (\hat\theta_\DF) \big] = \frac{1-f}{n}\frac{1}{N}\sum_{i=1}^N \mathbb{E}_P \big[ \mathbb{E}_P[ (Z_i - \hat{Z}_i)^2 \,|\, X] \big] 
    = \frac{1-f}{n}\mathbb{E}\big[\Var_P(Z \,|\, X) \big]. 
\end{align*}
The remaining part of the proof is straightforward and is thus omitted.

%% file: sections/extended_results.tex
\section{Extended Results and Analyses}
\label{app:additional_results}

\subsection{Detailed Analysis of Main Results in \Cref{sec:results}}\label{app:experiments_breakdownresults}

The datasets and tasks included in our experiments of \Cref{sec:results}, together with the efficiency of $\HT$ under simple random sampling relative to other methods, are listed in \Cref{table:results_summary}.

\begin{table}[H]
\centering
\caption{
    Breakdown of results in \Cref{fig:in_distribution_results}.
    Each number corresponds to the relative efficiency of model accuracy estimates obtained through different sampling designs and estimators compared to $\HT$ under $\SI$, the Horvitz-Thompson estimator under simple random sampling. 
    The proxy $\hat{Z}$ is constructed using model predictions $\hatm$, surrogate model predictions $\hatm^*$, and calibrated predictions of a surrogate model on in-distribution data $\hatm^{*c}$.
    Note that ``emb'' refers to the embeddings in the table. 
}
\label{table:results_summary}
\begin{adjustbox}{width=0.83\textwidth,height=0.95\textheight,keepaspectratio, center}
\setlength{\tabcolsep}{3pt}
\begin{tabular}{@{}lccc|cccc|ccc@{}}
\toprule
\multicolumn{1}{c} {\textbf{Datasets}} & \multicolumn{10}{c}{\textbf{Methods}} \\
\cmidrule(l){2-11} 
& \multicolumn{3}{c}{$\SI$ + $\DF$} & \multicolumn{4}{c}{$\ST, p$ + $\HT$} & \multicolumn{3}{c}{$\ST, o$ + $\HT$} \\
\cmidrule(lr){2-4} \cmidrule(lr){5-8} \cmidrule(lr){9-11}
\textbf{Dataset Name and Reference} & $\hatm$ & $\hatm^*$ & $\hatm^{*c}$ & emb & $\hatm$ & $\hatm^*$ & $\hatm^{*c}$ & $\hatm$ & $\hatm^*$ & $\hatm^{*c}$ \\
\midrule
Caltech 101 \cite{fei2004learning} & 0.66 & 0.66 & 0.54 & 0.75 & 0.57 & 0.60 & 0.54 & 0.47 & 0.81 & 0.33 \\ 
  Stanford Cars \cite{KrauseStarkDengFei-Fei_3DRR2013} & 0.69 & 0.35 & 0.34 & 0.98 & 0.69 & 0.34 & 0.33 & 0.51 & 0.24 & 0.23 \\ 
  CIFAR-10 \cite{krizhevsky2009learning} & 0.77 & 0.45 & 0.38 & 0.95 & 0.74 & 0.41 & 0.38 & 0.60 & 0.35 & 0.20 \\ 
  CIFAR-100 \cite{krizhevsky2009learning} & 0.69 & 0.50 & 0.46 & 0.92 & 0.67 & 0.47 & 0.46 & 0.75 & 0.52 & 0.37 \\ 
  CLEVR (distance)  \cite{johnson2017clevr} & 1.16 & 1.47 & 1.01 & 0.99 & 0.97 & 0.99 & 1.01 & 1.11 & 1.08 & 1.03 \\ 
  CLEVR (count)  \cite{johnson2017clevr} & 0.73 & 0.63 & 0.50 & 0.77 & 0.67 & 0.52 & 0.50 & 0.78 & 0.63 & 0.44 \\ 
  Describable Text Features \cite{cimpoi14describing} & 0.82 & 0.71 & 0.65 & 0.95 & 0.81 & 0.64 & 0.65 & 1.15 & 0.87 & 0.69 \\ 
  DR Detection \cite{diabetic-retinopathy-detection} & 1.02 & 1.11 & 0.97 & 0.96 & 1.00 & 0.97 & 0.97 & 1.04 & 1.03 & 0.95 \\ 
  DMLab Frames \cite{zhai2019visual} & 1.22 & 1.26 & 0.99 & 0.99 & 1.00 & 1.01 & 0.98 & 1.09 & 1.55 & 1.01 \\ 
  dSprites (orientation) \cite{dsprites17} & 1.06 & 1.21 & 0.95 & 0.93 & 0.98 & 1.02 & 0.95 & 1.13 & 1.05 & 0.89 \\ 
  dSprites (x position) \cite{dsprites17} & 1.03 & 1.04 & 0.97 & 1.00 & 1.02 & 1.01 & 0.97 & 1.08 & 1.24 & 1.01 \\ 
  dSprites (y position) \cite{dsprites17} & 1.12 & 1.82 & 0.99 & 1.01 & 0.97 & 1.00 & 0.99 & 1.03 & 1.01 & 0.96 \\ 
  EuroSAT \cite{helber2019eurosat} & 0.85 & 0.69 & 0.65 & 0.75 & 0.84 & 0.66 & 0.66 & 0.95 & 0.74 & 0.63 \\ 
  FGVC aircraft \cite{maji13fine-grained} & 0.83 & 0.76 & 0.71 & 0.91 & 0.80 & 0.68 & 0.70 & 0.80 & 0.73 & 0.63 \\ 
  Oxford 102 Flower \cite{nilsback2008automated} & 0.68 & 0.41 & 0.38 & 0.94 & 0.67 & 0.40 & 0.38 & 0.64 & 0.32 & 0.28 \\ 
  GTSRB \cite{stallkamp2011german} & 0.72 & 0.46 & 0.47 & 0.71 & 0.72 & 0.45 & 0.47 & 0.78 & 0.41 & 0.43 \\ 
  ImageNet-A \cite{hendrycks2021nae} & 1.06 & 0.81 & 0.60 & 0.99 & 0.95 & 0.59 & 0.60 & 1.38 & 0.82 & 0.50 \\ 
  ImageNet-R \cite{hendrycks2021many} & 0.63 & 0.31 & 0.30 & 0.96 & 0.61 & 0.30 & 0.30 & 0.58 & 0.28 & 0.21 \\ 
  ImageNet-1K \cite{imagenet15russakovsky} & 0.74 & 0.54 & 0.51 & 0.97 & 0.73 & 0.51 & 0.51 & 0.92 & 0.67 & 0.46 \\ 
  ImageNet Sketch \cite{wang2019learning} & 0.72 & 0.54 & 0.49 & 0.97 & 0.70 & 0.50 & 0.50 & 0.88 & 0.63 & 0.45 \\ 
  ImageNetV2 \cite{recht2019imagenet} & 0.75 & 0.58 & 0.52 & 0.97 & 0.74 & 0.54 & 0.52 & 0.97 & 0.70 & 0.50 \\ 
  KITTI Distance \cite{Geiger2013IJRR} & 1.06 & 1.12 & 0.92 & 0.69 & 0.97 & 0.93 & 0.88 & 1.09 & 1.04 & 0.95 \\ 
  MNIST \cite{deng2012mnist} & 0.73 & 0.27 & 0.19 & 0.47 & 0.66 & 0.18 & 0.19 & 0.67 & 0.11 & 0.12 \\ 
  ObjectNet \cite{barbu2019objectnet} & 0.75 & 0.51 & 0.41 & 0.96 & 0.72 & 0.45 & 0.41 & 1.06 & 0.60 & 0.35 \\ 
  Oxford-IIIT Pet \cite{parkhi2012cats} & 0.73 & 0.36 & 0.41 & 0.98 & 0.73 & 0.35 & 0.42 & 0.46 & 0.19 & 0.20 \\ 
  PASCAL VOC 2007 \cite{pascal-voc-2007} & 0.76 & 0.85 & 0.72 & 0.88 & 0.75 & 0.74 & 0.72 & 1.07 & 1.14 & 0.71 \\ 
  PCam \cite{veeling2018rotation} & 1.03 & 1.17 & 0.99 & 0.91 & 0.99 & 0.98 & 0.99 & 1.05 & 1.05 & 1.03 \\ 
  Rendered SST-2 \cite{socher2013recursive} & 1.06 & 1.14 & 0.98 & 0.98 & 1.00 & 0.98 & 0.98 & 1.11 & 1.15 & 1.02 \\ 
  NWPU-RESISC45 \cite{cheng2017remote} & 0.80 & 0.63 & 0.55 & 0.96 & 0.78 & 0.58 & 0.55 & 0.97 & 0.77 & 0.49 \\ 
  SmallNorb (Azimuth) \cite{lecun2004learning} & 0.97 & 1.24 & 1.06 & 1.03 & 0.93 & 0.98 & 1.05 & 0.99 & 1.14 & 1.11 \\ 
  smallNORB (Elevation) \cite{lecun2004learning} & 0.99 & 1.09 & 1.04 & 0.99 & 0.97 & 0.97 & 1.03 & 1.01 & 1.09 & 1.07 \\ 
  STL-10 \cite{coates2011analysis} & 0.75 & 0.31 & 0.34 & 0.95 & 0.71 & 0.29 & 0.33 & 0.43 & 0.19 & 0.23 \\ 
  SUN397 \cite{Xiao:2010} & 0.77 & 0.66 & 0.61 & 0.99 & 0.77 & 0.62 & 0.61 & 0.85 & 0.72 & 0.58 \\ 
  Street View House Numbers \cite{netzer2011reading} & 0.76 & 0.81 & 0.68 & 0.77 & 0.75 & 0.66 & 0.68 & 0.83 & 0.87 & 0.71 \\ 
\bottomrule
\end{tabular}
\end{adjustbox}
\end{table}

\subsection{Experiments on Other Classification Metrics}\label{app:experiments_othermetrics}

In the following suite of experiments, we consider the following evaluation metrics for the predictions $\hatm(X)=(\hatm_1(X), \dots, \hatm_K(X))$ made by the classifier $\hatm$: 
\begin{itemize}[leftmargin=7mm]
    \item Mean squared error (MSE), where $Z=(1 - \hatm_Y(X))^2$. The expected value of $Z$ given $X$ is $\mathbb{E}_P[Z \,|\, X] = \sum_{k=1}^K \mathbb{P}_P(Y=k \,|\, X)(1 - \hatm_k(X))^2$.
    \item Cross-entropy loss, where $Z = - \log \hatm_Y(X)$. The expected value of $Z$ given $X$ is $\mathbb{E}[Z \,|\, X] = \smash{-\sum_{k=1}^K \mathbb{P}_P(Y=k \,|\, X) \log \hatm_k(X)}$.
\end{itemize}

To estimate $S^2_{Z_h}$, which is needed for allocating the budget to strata under Neyman allocation (i.e., set $n_h$), we use the plug-in estimator $\smash{\hat{S}^2_{Z_h} = \frac{1}{N_h} \sum_{i\in \cD_h} \hat{Z}^{(2)}_i - \hat{\bar{Z}}_{\cD_h}^{(2)}}$ where $\hat{Z}^{(2)}_i$ is an estimator of $\mathbb{E}[Z_i^2 \,|\, X_i]$ and $\hat{\bar{Z}}_{\cD_h}^{(2)}$ is its empirical average taken over $\cD_h$.

\begin{figure*}[!ht]
    \centering
    \includegraphics[width=\textwidth]{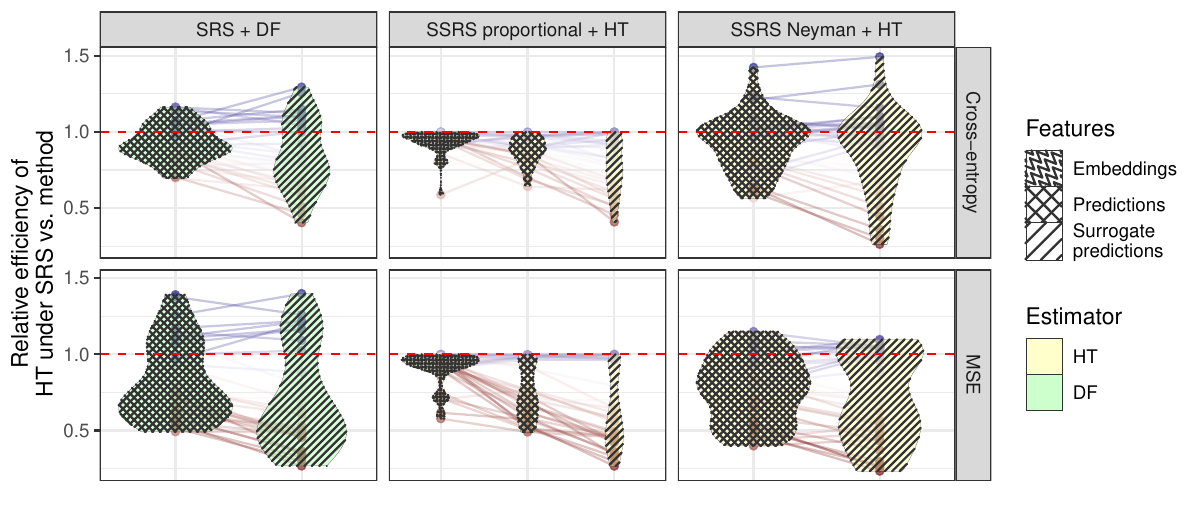}
    \caption{
        \textbf{Comparison of efficiency across stratification procedures, sampling designs, and estimators for estimating $\mse$ and cross-entropy.}
        We evaluate the zero-shot accuracy of CLIP ViT-B/32 and generate surrogate predictions using CLIP ViT-L/14, also in the zero-shot setting.
        For more details, refer to \Cref{fig:in_distribution_results}.
    }
    \label{fig:other_metrics_results}
\end{figure*}

\Cref{fig:other_metrics_results} shows the results obtained using the same setup and models as in \Cref{sec:results}. Similar to the accuracy analysis, we observe that the proportional allocation estimates made by stratifying over $\auxmod$ using ViT-L/14's predictions generally outperform those using CLIP ViT-B/32's predictions, which in turn are more precise than those made by stratifying on CLIP ViT-B/32 embeddings. 
Estimates from proportional allocation are more accurate than those from Neyman allocation on some tasks where Neyman sometimes underperforms compared to the baseline. However, proportional allocation does not achieve the substantial improvements seen with Neyman's on other tasks. The $\DF$ estimator performs better than the baseline on some tasks but worse on others. In additional experiments we found that using a $\auxmod$ that is trained on in-distribution validation data boosts the performance of both $\DF$ and Neyman, allowing them to always improve upon the baseline. This is consistent with the findings in \Cref{fig:in_distribution_results}, where the lack of calibration in predictions can lead to larger variances compared to the baseline. Overall, each method significantly reduces the error in estimating model $\mse$ and cross-entropy loss.

\clearpage

\subsection{Experiments on CLIP Models with Linear Probing}\label{sec:experiments_lp}

We compare the efficiency of the methods in estimating the binary classification accuracy of predictions made by CLIP ViT-B/32 and CLIP ViT-L/14 using linear probing. 
In this setup, the model embeddings are frozen and a single linear layer is trained on top of them. 
We train it using the LAION CLIP repository code with the default data splits \cite{laion_clip_benchmark}. 
Across the tasks evaluated in the zero-shot setting and with linear probing, the latter consistently achieves higher accuracy compared to the zero-shot setting.

\begin{figure*}[!ht]
    \centering
    \includegraphics[width=\textwidth]{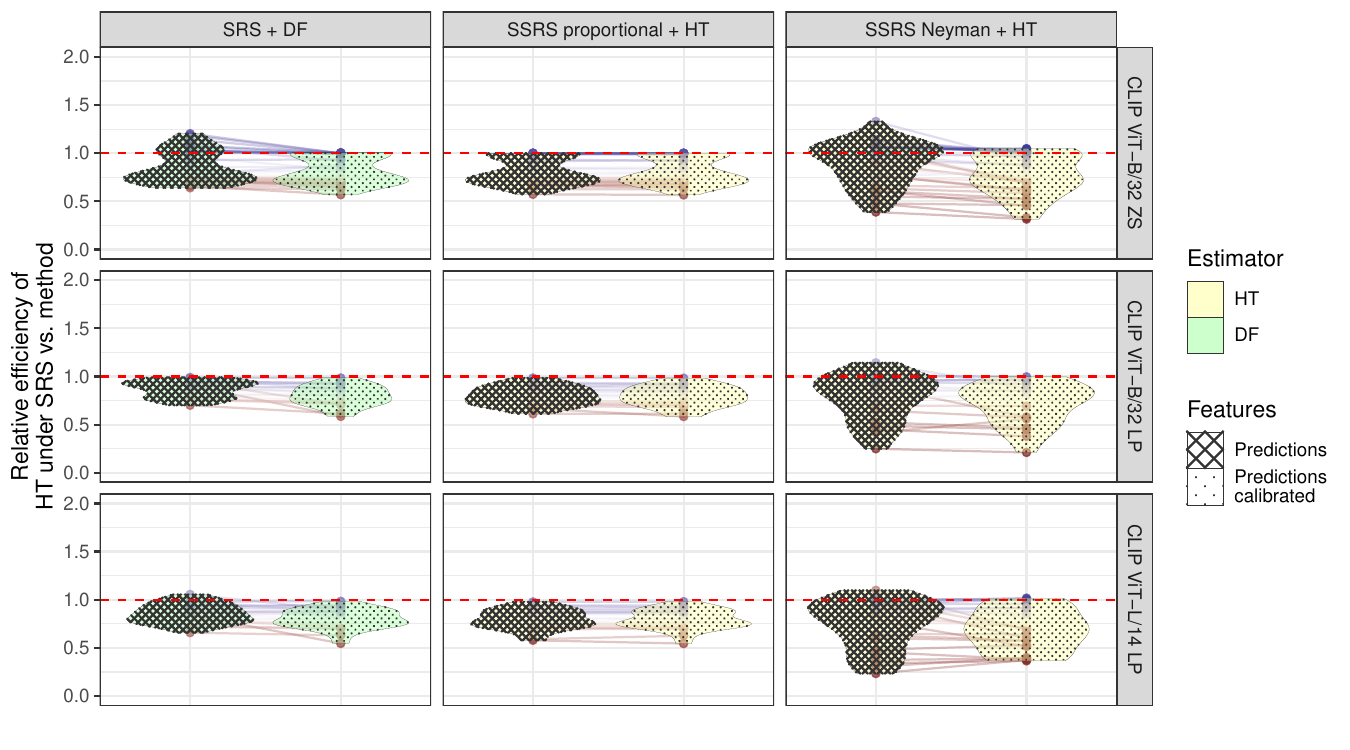}
    \caption{
        \textbf{Comparison of efficiency across sampling designs, estimators, and CLIP models in the zero-shot setting (ZS) and with linear probing (LP).} 
        In this figure, we present the results specifically for the proxy $\hat{Z}$ of $Z$ built on the model being evaluated. 
        For a more detailed explanation of the figure, please see \Cref{fig:in_distribution_results}.
    }
    \label{fig:linear_probing_results}
\end{figure*}

The main results from this set of experiments are shown in \Cref{fig:linear_probing_results}.
For easy comparison, we also report the efficiency of the methods for CLIP ViT-B/32 in the zero-shot setting. 
In contrast to \Cref{fig:in_distribution_results}, we observe that with linear probing, most of the methods outperform the baseline of $\HT$ under $\SI$. 
However, in the zero-shot setting, the methods tend to perform worse. 
This could be attributed to the lower $\mse$ achieved by training the linear layer, as discussed in \Cref{sec:methods}.
Consistent with our findings in \Cref{sec:results}, calibration of the proxy $\hat{Z}$ (based only on $\hatm$) improves efficiency for $\HT$ under Neyman allocation and for $\DF$ under $\ST$, but not for $\HT$ under proportional allocation.
In addition, the differences in efficiency between ViT-B/32 and ViT-L/14 with linear probing become less pronounced compared to the zero-shot setting.

\clearpage

\subsection{Experiments on Other Visual Encoders of CLIP Models}\label{app:experiment_otherarchitectures}

We compare the methods in estimating the zero-shot classification accuracy of CLIP models with ResNet 50 \cite{he2016deep} and ConvNeXT base \cite{liu2022convnet} as visual encoders.
We obtain the surrogate predictions using ResNet 101 and ConvNeXT XXLarge respectively. 

\begin{figure*}[!ht]
    \centering
    \includegraphics[width=\textwidth]{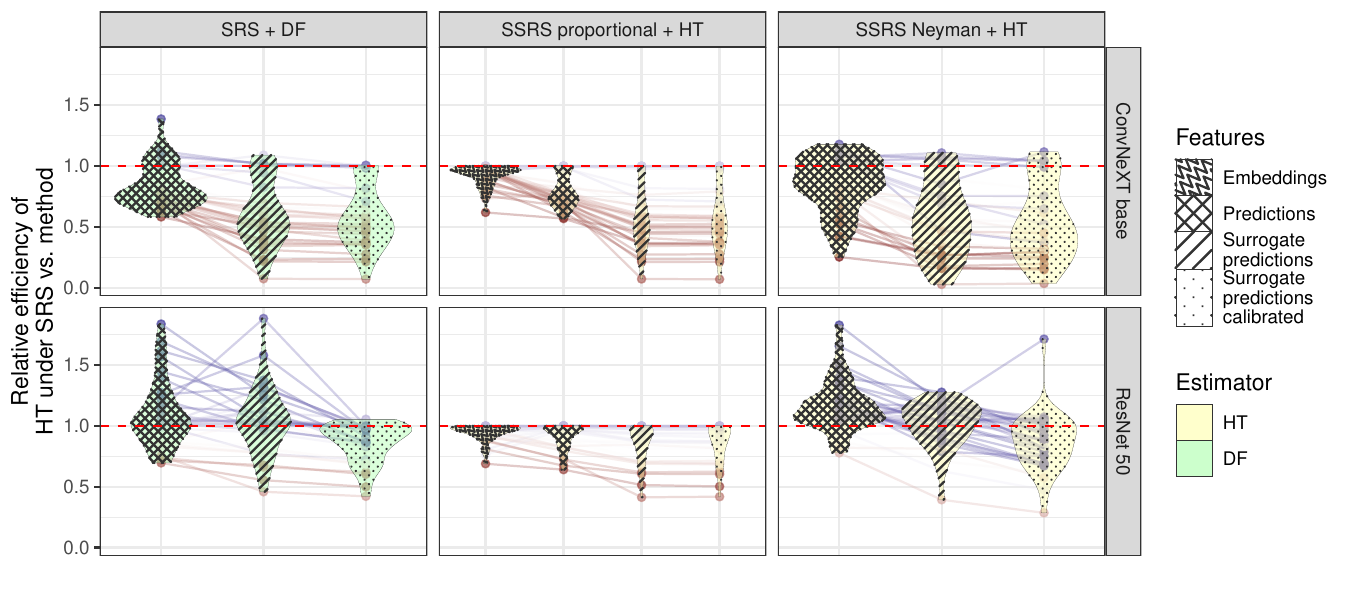}
    \caption{
        \textbf{Comparison of efficiency across sampling designs, estimators, and models.}
        We evaluate the performance of ResNet 50 on the LAION CLIP benchmark tasks using surrogate predictions from ResNet 101. 
        Both models are pretrained on the same data. 
        Similarly, for ConvNeXT, we assess the accuracy of the base model using surrogate predictions from ConvNeXT XXLarge. 
        Please see \Cref{fig:in_distribution_results} for a detailed explanation of the elements in the figure.
    } 
    \label{fig:other_visual_encoders}
\end{figure*}

The results are shown in \Cref{fig:other_visual_encoders}. At a high level, the takeaways in \Cref{sec:results} hold in this context as well. More specifically, using $\ST$ with proportional allocation always lowers the variance of the estimates of model accuracy compared to using $\HT$ under $\SI$. The stratification based on the predictions is more effective than the one on the embeddings. Similarly to \Cref{fig:in_distribution_results}, the efficiency of $\DF$ under $\SI$ and of $\HT$ under Neyman allocation varies across datasets and is not always superior to the baseline. Calibration, however, improves efficiency across most datasets. 
As noted previously, we also find that leveraging surrogate predictions from models with higher accuracy typically enhances the precision of our estimates for these architectures as well. 
Lastly, we find that ConvNeXT achieves far higher performance in the classification tasks compared to ResNet and the efficiency gains over $\HT$ under $\SI$ for the former are consistently larger across all methods. 

\clearpage
\subsection{Comparison of In- versus Out-of-Distribution Data}
\label{app:experiment_ood}

\begin{figure}[!ht]
    \centering
    \includegraphics[width=\textwidth]{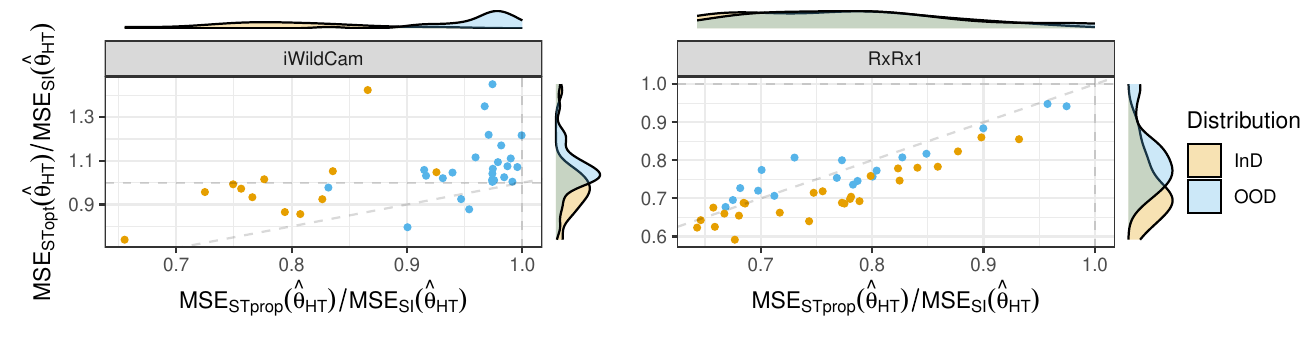}
    \caption
    {
    \textbf{Comparison of the efficiency of sampling designs and estimators on in-distribution versus out-of-distribution data.} 
    The relative efficiencies of $\HT$ under $\SI$ vs. the $\HT$ estimator under $\ST$ with proportional allocation (horizontal axis) and Neyman allocation (vertical axis) are shown in the plot. 
    The methods estimate the classification accuracy of a Resnet 18 model trained and evaluated on the WILDS-iWildCam and WILDS-RxRx1 datasets. Stratification is done on $\auxmod$ using the predictions of $Z$ made by the models. 
    Each point in the plot represents one domain in the datasets, with kernel density estimates of these points shown on the margins.
    We observe that the methods perform better compared to the baseline when the model is evaluated on in-distribution data. 
    On the out-of-distribution data of the iWildCam dataset, Neyman allocation generally performs worse than proportional allocation and often worse than $\SI$.
    }
    \label{fig:indvsood}
\end{figure}

To evaluate the performance of our methods on in- vs. out-of-distribution data, we finetune a ResNet 18 model on the RxRx1 \cite{taylor2019rxrx1} and iWildCam \cite{beery2020iwildcam} datasets from the WILDS out-of-distribution benchmark \cite{koh2021wilds}. 
This is done using SGD on the official train splits of the datasets. We then calibrate the models using the in-distribution validation split, and evaluate their performance on in- and out-of-distribution test domains.
In \Cref{fig:indvsood}, we compare the performance of the in-distribution and out-of-distribution settings.  
The figure highlights that when estimating model performance, efficiency gains from stratified sampling procedures are likely to be higher on the in-distribution data.

